\documentclass[11pt,a4paper]{article}
\usepackage{authblk}
\usepackage[hyperref]{acl2021}
\usepackage{times}
\usepackage{multirow}
\usepackage{graphics}
\usepackage{latexsym}
\usepackage{layouts}
\usepackage{graphicx}
\usepackage{tabularx}
\graphicspath{ {./images/} }
\usepackage{tikz}
\usetikzlibrary{shapes.geometric, arrows, positioning}
\usepackage{lipsum,adjustbox}
\usepackage{array}
\newcolumntype{L}[1]{>{\raggedright\let\newline\\\arraybackslash\hspace{0pt}}m{#1}}
\newcolumntype{C}[1]{>{\centering\let\newline\\\arraybackslash\hspace{0pt}}m{#1}}

\usepackage{microtype}

\newcounter{ex}
\newenvironment{expl}
{\begin{enumerate} \refstepcounter{ex} \item[(\theex)]}
{\end{enumerate}
}

\aclfinalcopy % Uncomment this line for the final submission
\title{UIUC\_BioNLP at SemEval-2021 Task 11: \\A Cascade of Neural Models for Structuring Scholarly NLP Contributions}

\author{Haoyang Liu, Janina Sarol, \and Halil Kilicoglu\\
  School of Information Sciences, University of Illinois at Urbana-Champaign \\
  \texttt{\{hl57, mjsarol, halil\}@illinois.edu} \\}
\date{}
\begin{document}
\maketitle
\begin{abstract}
We propose a cascade of neural models that performs sentence classification, phrase recognition, and triple extraction to automatically structure the scholarly contributions of NLP publications in English. To identify the most important contribution sentences in a paper, we used a BERT-based classifier with positional features (Subtask 1). A BERT-CRF model was used to recognize and characterize relevant phrases in contribution sentences (Subtask 2). We categorized the triples into several types based on whether and how their elements were expressed in text, and addressed each type using separate BERT-based classifiers as well as rules (Subtask 3). Our system was officially ranked second in Phase 1 evaluation and first in both parts of Phase 2 evaluation. After fixing a submission error in Phase 1, our approach yielded the best results overall. In this paper, in addition to a system description, we also provide further analysis of our results, highlighting its strengths and limitations. We make our code publicly available at \url{https://github.com/Liu-Hy/nlp-contrib-graph}.
%For sentence classification, a model is customized to incorporate useful features. For triple extraction, triples are categorized into different types and predicted separately with proper manipulation of the input. In the end-to-end evaluation, the performance of our system on triple extraction is comparable to intra-annotator agreement, showing the effectiveness of our approach. Finally, analysis is done on the strengths and limitations of our system. This paper describes the best performing system for the SemEval-2021 NLP Contribution Graph task. 
\end{abstract}

\section{Introduction}
With the deluge of scientific publications in recent years, keeping pace with the literature and managing information overload have become increasingly challenging for researchers. There is a growing need for tools that can automatically extract and structure semantic information from scientific publications to facilitate advanced approaches to information access and knowledge curation~\citep{shen-etal-2018-web}.  

The field of natural language processing (NLP) has witnessed an enormous growth in recent years with advances in deep learning, and there are increasing efforts in developing methods to extract scholarly knowledge from NLP publications~\citep{QasemiZadeh-Schumann-2016-acl,dsouza2020nlpcontributions}. One such effort is \textsc{NlpContributions}, an annotation scheme for describing the scholarly contributions in NLP publications and a corpus annotated using this annotation scheme~\cite{dsouza2020nlpcontributions}. This corpus has been proposed for training and testing of machine reading models, whose output can be integrated with the Open Research Knowledge Graph framework (ORKG)~\citep{10.1145/3360901.3364435}. ORKG formalizes the research contributions of a scholarly publication as a knowledge graph, which can further be linked to other publications via the graph. The goal of the NLPContributionGraph (NCG) shared task ~\cite{ncg} is to facilitate the development of machine reading models that can extract ORKG-compatible scholarly contribution information from NLP publications. The shared task consists of three subtasks (see~\citet{ncg} for a more detailed description):
\begin{itemize}
    \item Subtask 1: Identification of contribution sentences from NLP publications
    \item Subtask 2: Recognition of scientific terms and relations in contribution sentences
    \item Subtask 3: Extraction and classification of triples that pair scientific terms with relations
\end{itemize}
\begin{figure*}%[!htbp]
\begin{center}
\includegraphics[width=\textwidth]{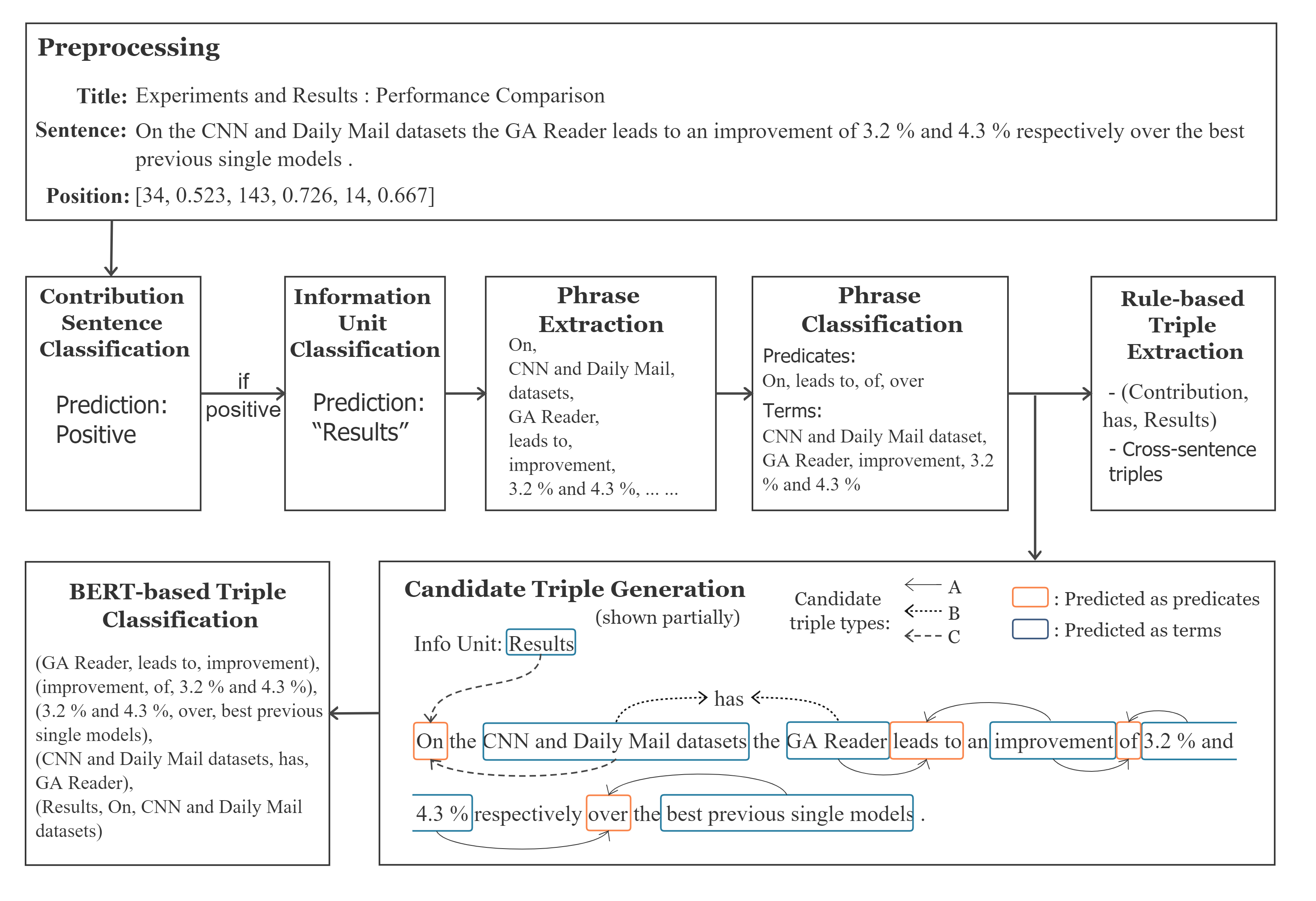}
\caption{\label{dia}End-to-end system diagram.}
\end{center}
\end{figure*}
In this paper, we describe our contribution to NCG shared task. 
%We completed all three subtasks, and participated in all phases of evaluation. 
We built a cascade of neural classification and sequence labeling models based on BERT~\citep{devlin-etal-2019-bert}. For subtask 3, we characterized triples based on whether and how their elements are expressed in text, and employed different models for each category. We also explored rule-based heuristics to improve model performance. Our models had the best overall performance in the shared task (57.27\%, 46.41\%, and 22.28\% F$_{1}$ score in subtasks 1, 2, and 3, respectively). The results are encouraging for extracting scholarly contributions from scientific publications, although there is much room for improvement.

\section{System Overview} \label{method}
In this section, we first describe our data preprocessing steps. Next, we discuss our models for each subtask, and the experimental setup for our end-to-end system (Phase 1). We provide an overview of the system in Figure~\ref{dia} and provide examples for illustration, when necessary.

\subsection{Data preprocessing} \label{pre}
The participants of the shared task were provided three kinds of input: a) plain text files of the publications converted from PDF using Grobid\footnote{\url{https://grobid.readthedocs.io/}}, b) sentences and tokens identified using Stanza~\citep{qi-etal-2020-stanza}, and c) triples and source texts organized by their information units (e.g., \textsc{Approach}) in JSON format. 

\subsubsection{Identifying headers and positional information}
One major preprocessing step was to identify section headers in the publications and associate them with individual sentences. For sentence classification (subtask 1), we incorporated the topmost and innermost section headers associated with a sentence into its representation. The topmost header indicates the general role that a sentence plays in the article, while the innermost header provides more specific context for the sentence. For example, one topmost/innermost header pair is \textsc{Experiment}/\textsc{Data Set and Experiment Settings}. 

%Our discussion involves the following three kinds of files provided by the task organizers.
%\paragraph{Grobid files} These files contain the plain text converted from each of the NLP papers using the Grobid parser~\citep{GROBID}.
%\paragraph{Stanza files} These files contain the output of tokenization and sentence segmentation on Grobid files using Stanza~\citep{qi-etal-2020-stanza}, where every heading and ordinary sentence occupies a line. 
%\paragraph{JSON files} In the annotation of each paper, the triples are grouped under different information units, and organized as a semantic tree for each unit in JSON format. The source text of each triple, which consists of one or more sentences, is included in the tree as the leaf node. 

In the absence of explicit section information in the input, we used rule-based heuristics to extract these headers. With the first heuristic (Heuristic1), we simply identified the sentences following blank lines in plain text files as section headers. In Heuristic2, we first identified candidate headers as sentences that contain fewer than 10 words, have the first letter capitalized, do not end with several stopwords (\emph{by}, \emph{as}, \emph{in}, \emph{that}, or \emph{and}), do not contain question marks in the middle or end with some punctuation (comma, colon or full stop). Next, we determined the case format used for headers in the publication by counting the occurrences of each case format type (e.g., all uppercase: EXPERIMENTAL SETUP). Among the headers that conform to the determined case format, we distinguished topmost headers as those that contain several lexical cues (e.g., \emph{background}, \emph{method}) and are shorter than 5 words. Finally, we associated each sentence with the nearest preceding topmost and innermost header. 

To incorporate headers into the sentence representation, we join the topmost and innermost header together with a colon between them and refer to it as the ``title'' of the sentence. In the case where a sentence is directly governed by a top-level header or it is a header itself, the title consists of the topmost header only.

We characterize the position of each sentence in the document with a combination of numeric features: 
\begin{itemize}
\item The offset of the sentence in the entire paper.
\item The offset of the sentence with respect to its topmost header.
\item The offset of the sentence with respect to the header extracted using Heuristic1. 
\end{itemize}
Each of these offset features are divided by the number of sentences 
in the corresponding discourse (entire paper or the section) to extract a proportional sentence position feature. Thus, for every sentence, a total of six positional features (three offsets, three proportional sentence positions) are computed. 
%We calculate the 6 position features of every sentence when preprocessing the data.

\subsubsection{JSON Parsing} \label{json}
We created two additional models to assist with triple extraction: a) a multi-class sentence classifier that labels each sentence with a single information unit and b) a binary phrase classifier that labels phrases as scientific terms vs. predicates (described below). To train these models, we extracted additional information from JSON files. 
%(In the JSON file of each info unit, the source text from which the triples are found is annotated as the leaf node of the JSON tree.)
%First, we associated contribution sentences with information units that characterize the triples in the sentence.
First, we matched the contribution sentences with the source text in the JSON files to get the information unit labels of the sentences. Second, we aligned the phrases with the triples in the same information unit, and determined whether each phrase is a predicate or term based on its location in the triple. 

\subsection{Subtask 1: Contribution Sentence Classification}
We built a binary classifier to determine whether each sentence describes a contribution of the publication. Our analysis revealed that this decision was not simply based on the semantics of the sentence, but also its position in the document. On one hand, the section header associated with the sentence provides important clues about the role of the sentence in the larger context. For example, the header ``Related Work" indicates that sentences in this section are likely to discuss the contributions of prior research. On the other hand, some parts of the documents are naturally more salient than others (e.g. title, abstract, the first few lines of each section), where authors tend to summarize the most important information. To operationalize these insights, we designed a model that captures the information about the sentence, its topmost and innermost headers as well as its position in the document, as discussed above.
%To achieve this, we did thorough analysis on the data to develop rules for heading detection, and find useful features to represent the sentence position.

\begin{figure}[!ht]
\begin{center} 
\includegraphics[width=\columnwidth]{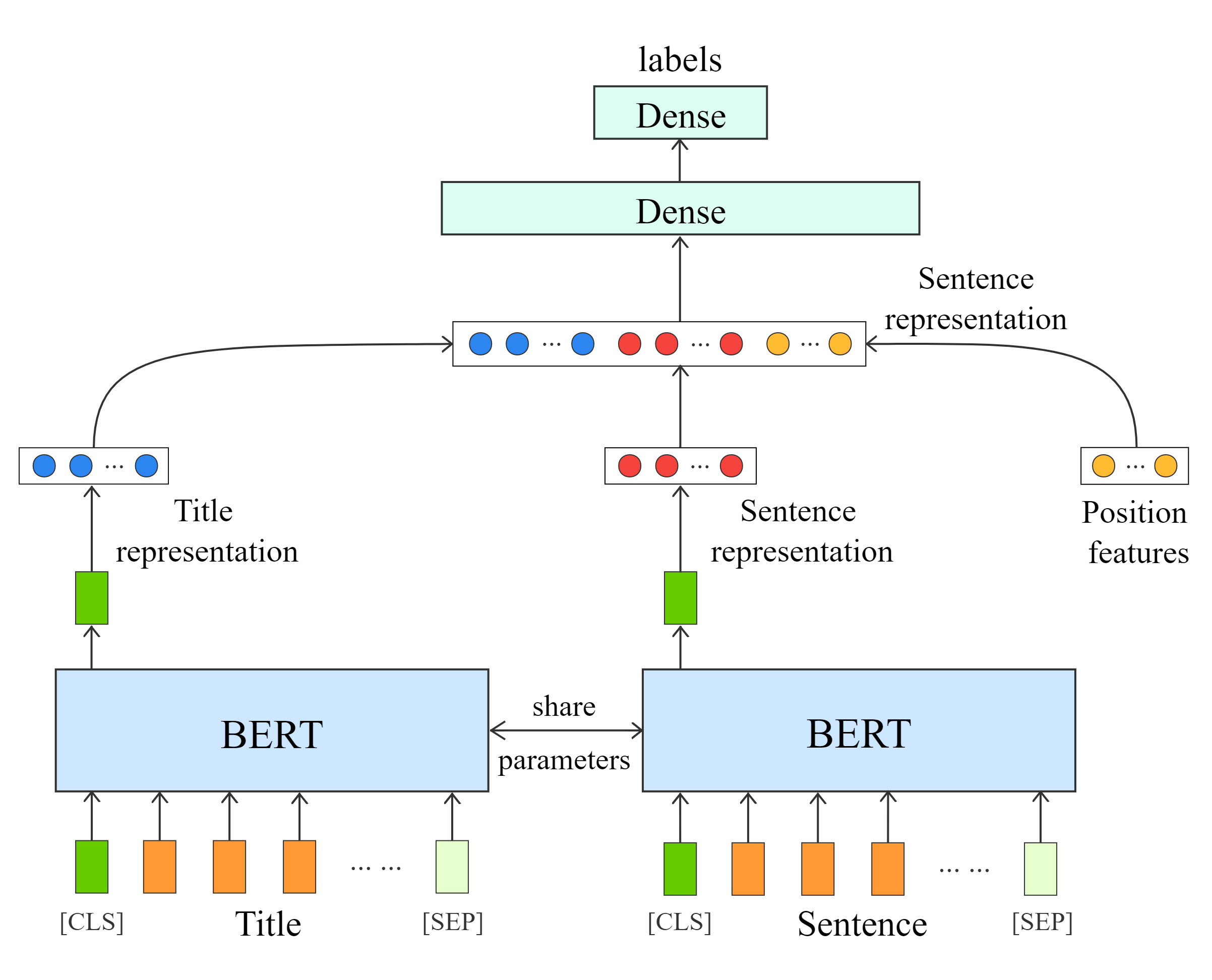}
\caption{\small Sentence classification model architecture\label{model}} 
\end{center} 
\end{figure}

We used a BERT model to encode the sentence and its title (i.e., concatenated headers) separately and concatenated their textual representation together with the positional features to obtain a sentence representation. We then fed this representation into two dense layers, and used a final softmax layer for classification (Figure~\ref{model}).
\subsection{Subtask 2: Phrase Recognition} \label{ner}
Subtask 2 is similar to a named entity recognition (NER) task, although the participating systems were only required to extract relevant text spans and not to categorize them.
%Named Entity Recognition(NER) task in that it involves extraction of text spans. 
One major difficulty with this subtask is that phrases do not align neatly with sentence constituents (e.g., noun phrases) and they vary greatly in length and in what counts as their boundaries (e.g. \emph{best results} and \emph{our best results} are both valid phrases).

For this subtask, we used a BERT-CRF model for phrase extraction and type classification~\cite{souza19}. The raw text of the sentence is taken as the model input. A BIO scheme that incorporates phrase types (scientific term vs. predicate) is used (e.g., B-Predicate, I-Term, O). The probabilities produced by the BERT model are fed into a Conditional Random Field (CRF) layer~\cite{crf} for end-to-end training. We note that while phrase type classification is not necessary for subtask 2, we perform it since it is useful for our subtask 3 model, described next.

\subsection{Subtask 3: Triple Extraction} \label{trip}
Subtask 3 involves organizing phrases into triples. In information extraction, semantic triples are typically composed of subject, predicate, and object terms each corresponding to specific textual spans. This is not always the case in this subtask. While in most cases all three terms are extracted from a single sentence, a non-negligible number of triples consist of at least one phrase that does not come from the sentence (e.g. (\textsc{Tasks}, \texttt{has}, \emph{Coreference resolution}), where the subject is an information unit and the predicate is not a sentence element).

To better understand triple characteristics, we categorized them into several types based on their composition, and created separate relation classification models for each type. The triple categorization is presented in Table~\ref{tripletypes}. For each type, we list their functions in information organization, their proportion to all triples, along with some examples. We note that input to the training process for triple extraction varies by the type of the triple (described for each type in Section~\ref{neural_triple}).

\begin{table*}[!h]%!htbp
\centering
\fontsize{10.5}{11}\selectfont
\begin{tabular}{|C{1.4cm}|L{4.5cm}|L{4cm}|L{2.5cm}|C{1cm}|}
\hline
& Composition & Examples & Role & Pct. \\
\hline
Type A & Three phrases in a sentence & (\emph{Deep - ED}, \texttt{obtain}, \emph{BLEU score})%, (BLEU score, \texttt{of}, 36.3)
& Organize the semantics of a sentence. & 57\% \\ \hline
Type B & Two terms in a sentence with an added predicate \texttt{has} or \texttt{name} & (\emph{ByteNet Decoder}, \texttt{has}, \emph{30 residual blocks})%, (external knowledge sources, name, Wikipedia) 
& Organize the semantics of a sentence. & 7\% \\ \hline
Type C & Information unit (subject), and two phrases in a sentence (predicate and object)  & (\textsc{Hyperparameters}, \texttt{use}, \emph{cross - entropy loss}) & Link a sentence to its information unit. & 9\% \\ \hline
Type D & Information unit (subject), \texttt{has} (predicate), and a term in the sentence (object) & (\textsc{Hyperparameters}, \texttt{has}, \emph{starting learning rate}) & Link a sentence to its information unit. & 9\% \\ \hline \hline
Type E & \textsc{Contribution} (subject), \texttt{has} (predicate), information unit (object) OR 
\textsc{Contribution} (subject), fixed (predicate), and a phrase (object) for the information units \textsc{Research problem} and \textsc{Code} & (\textsc{Contribution}, \texttt{has}, \textsc{Results}), (\textsc{Contribution}, \texttt{has research problem}, \emph{neural machine translation}) & Link the “Contribution” node of each paper to an information unit. & 9\% \\ \hline
Type F & Cross-sentence triples & (\emph{Positional Encoding}, \texttt{inject}, \emph{some information}) & Structure the information across sentences & 3\% \\ \hline
\end{tabular}
\caption{\label{tripletypes}
Triple types, their roles, and frequency. Types A-D are addressed using neural models and Types E-F with rules. 6\% of triples do not fit in these categories and are not shown.
}
\end{table*}

\subsubsection{Information Unit Classification} \label{infounit}
To aid triple extraction, we modified the binary classification model that we trained for subtask 1 to further classsify contribution sentences by their information units (multi-class classification). The process of labeling contribution sentences with information units was briefly described in Subsection~\ref{json}.  %After changing the output dimension, the above-mentioned model can also be used to classify each positive sentence into an information unit.

In analyzing the information units, we identified two special pairs (\textsc{Model} vs. \textsc{Approach} and \textsc{Experimental-setup} vs. \textsc{Hyperparameters}). In the dataset, no document contains both units of a pair. The decision of which unit to choose is made at the document level. Therefore, we merged the labels of similar units before feeding the examples into the multi-class classification model. 
 
After classification, we used lexical rules to split these units. Our rules were based on the following observations. First, the \textsc{Model} vs. \textsc{Approach} distinction seems related to how the authors mention their work in the abstract and section headers of the paper. Second, \textsc{Experimental-setup} is often used instead of \textsc{Hyperparameters} when the hardware or the framework used in the study is specified (e.g. \emph{V100 GPU, Pytorch}).

We did not recognize \textsc{Code} information units using this model, since we found that such sentences can be identified with a very high accuracy using a simple rule based on presence of a URL in the sentence. 

\subsubsection{Neural models for triple extraction} \label{neural_triple}
We extract triples of type A, B, C and D~(Table \ref{tripletypes}) by formulating them as neural relation classification tasks. All the classifiers are vanilla BERT classifiers (one linear layer followed by softmax). For each type, we observed the patterns in the training data, and addressed the most common ones. Ignoring the less frequent patterns inevitably led to a lower recall ceiling in our models. 

\paragraph{Type A} \label{typea}
This type, in which all triple elements are mentions in the sentence,  represents the majority of the triples. The corresponding model classifies the triples as a whole (``triple classification''). To the best of our knowledge, little research has been done on relation classification among three phrases; however, the Transformer model at the core of BERT is versatile enough to succeed in a wide range of tasks. As our training examples, we take every combination of a predicate and two terms in a sentence as a candidate triple, and train a model that predicts whether the three phrases constitute a triple or not. We encode the relation between three phrases by marking their boundaries in the sentence, as shown in Example~\ref{ex1}. We use angle brackets to enclose predicates, and square brackets to enclose terms.
\begin{expl}
\emph{In this paper , we explore an alternate [[ semisupervised approach ]] which does $<<$ not require $>>$ [[ additional labeled data ]] .}
\label{ex1}
\end{expl}
\paragraph{Type B} To identify triples of type B (two terms from the sentence and the relation type one of \texttt{has}, \texttt{name}, or \texttt{None}), we classify the relation between each pair of terms in a sentence that are not related by a type A triple.  
We found that 96\% of these triples preserve the order of the two terms in the sentence, so we also preserve the order for extraction.
\paragraph{Type C} 
Type C triples involve an information unit name as the subject along with a predicate and object from the sentence. We found that 89\% of these triples take the first predicate and the first term in a sentence as their predicate and object respectively. Furthermore, in 98\% of these sentences, the first predicate precedes the first term. Therefore, we classify each sentence whose first predicate precedes the first term, to predict whether a triple of this type can be extracted from the sentence. To train this classifier, we prepend the information unit name to the sentence text with a colon in between, as in Example~\ref{ex2} (\emph{Model} is the information unit).
\begin{expl}
\emph{[[ Model ]] : In this work , we $<<$ introduce $>>$ [[ a new type of linear connections ]] for multi - layer recurrent networks .}
\label{ex2}
\end{expl}
\paragraph{Type D} Type D triples are similar to Type C, but instead of a predicate phrase from the sentence, they involve the non-sentence predicate \texttt{has}. We found that 95\% of these triples in the training set take the first term in the sentence as their object, and the first predicate in the sentence, if one exists, almost always follows the first term. Therefore, we classify each sentence that conforms to this pattern, to predict whether the information unit name and the first term constitute a \texttt{has} relation. We prepend the info unit name to the sentence in the same way as in Type C.

\subsubsection{Rule-based triple extraction}
Triples of type E and F are extracted using heuristic rules. For type E, the subject is always \textsc{Contribution}. The predicate can be \texttt{has}, in which case the object is the name of an information unit. If the related information unit is \textsc{Code} or \textsc{Research problem}, the predicate is a fixed predicate (\texttt{Code} or \texttt{has research problem}, respectively) and the object is a phrase from the sentence. These rules use phrase and information units identified in earlier steps (Sections~\ref{ner} and~\ref{infounit}, respectively).

%These triples are extracted based on predictions on information units and phrases \textcolor{red}{BE MORE SPECIFIC??} 
%\textcolor{blue}{For every info unit, there must be a type E triple. If the unit is Code or Research problem, it uses the fixed predicate “Code” or “Research problem”, and take each term in the sentence as the object, to form a triple. Otherwise, it is (Contribution, has, unit name)} 

\begin{table*}[!htbp]
\centering
\resizebox{\textwidth}{!}{%
\begin{tabular}{|c|c|ccc|ccc|ccc|ccc|}
\hline
\multirow{2}{*}{} & \multirow{2}{*}{Avg F$_{1}$} & \multicolumn{3}{c|}{Information Units} & 
\multicolumn{3}{c|}{Sentences} & \multicolumn{3}{c|}{Phrases} &
\multicolumn{3}{c|}{Triples} \\
\cline{3-14}
& & F$_{1}$ & P & R & F$_{1}$ & P & R & F$_{1}$ & P & R & F$_{1}$ & P & R \\ \hline
Our system & 49.72 & 72.93 & 66.67 & 80.49 & 57.27 & 53.61 & 61.46 & \textbf{46.41} & 42.69 & 50.83 & \textbf{22.28} & 22.30 & 22.26 \\
\hline
IAA & 52.82 & \textbf{79.73} & 78.83 & 80.65 & \textbf{67.44} & 67.25 & 67.63 & 41.84 & 45.36 & 38.83 & \textbf{22.28} & 23.76 & 20.97 \\
\hline
\end{tabular}
}
\caption{\label{phase1}
End-to-end performance (Evaluation Phase 1). IAA: intra-annotator agreement.
}
\end{table*}

We developed the following rules to extract cross-sentence triples (type F):
\begin{enumerate}
    \item If the first sentence has a single entity, and the second sentence has at least 2 entities, we assign the entity in sentence 1 as the subject and the first and second entities in sentence 2 as the predicate and object, respectively. We add this triple to the list only if both subject and predicate are noun phrases, which prevents many false positives.
    %, particularly when the second entity is a predicate. 
    We also add the corresponding triple in the form of \textsc{info-unit}-\texttt{has}-\emph{subject} (e.g. \textsc{Model}-\texttt{has}-\emph{Encoder}). In many sentences that follow this rule, the first sentence is a section header. 
    
%    Sentence 1: \textit{Dynamic Routing}
    
%    Sentence 2: \textit{The basic idea of dynamic routing is to construct a non-linear map in an iterative manner ensuring that the output of each capsule gets sent to an appropriate parent in the subsequent layer}
    
%    Triple: \textit{(Dynamic Routing, to construct, non-linear map)}
    
    \item If the two sentences each contain a single term and sentence 1 term is a substring of sentence 2 term or if sentence 1 term is an acronym of sentence 2 term, we create the following triple: \emph{term 1}-\texttt{name}-\emph{term 2}. We extract a term's acronym by combining the initials of each token in the entity. An example of a term pair that follows this rule is (\emph{GLUE}, \emph{General Language Understanding Evaluation}). %\textcolor{red}{HOW DO WE KNOW ACRONYMS? IF THERE ARE RULES, SPECIFY}
\end{enumerate}

% [48, 4, 'Encoder :']
% [49, 5, 'The encoder is composed of a stack of N = 6 identical layers .']

These rules are applied to consecutive sentences only. In the training set, we found 812 triples that follow these rules, 649 (80\%) of which could be identified correctly using these rules.

\subsection{Experimental Setup}
We implemented our models using Simple Transformers\footnote{\url{https://github.com/ThilinaRajapakse/simpletransformers}}. We used SciBERT~\citep{beltagy-etal-2019-scibert} as the pre-trained language model. To train our models, we used a batch size of 16, and empirically found the best learning rate for each model between $10^{-5}$ and $10^{-4}$. One exception was that in our sentence classification model (subtask 1), we used a fixed learning rate of $10^{-5}$ to fine-tune the BERT, and a larger learning rate between $5\times10^{-5}$ and $10^{-3}$ for the dense layers. We used the AdamW optimizer~\cite{loshchilov17} and the polynomial decay scheduler with the power of 0.5. We ran the experiments on a Google Cloud VM instance, using a Tesla V100 GPU. 

\section{Results}
All the subtasks were evaluated on F$_{1}$ scores, and among them, triple extraction is evaluated by the micro-average of F$_{1}$ scores on each information unit. In the end-to-end evaluation (Phase 1), the participants were provided with the raw input to perform all three subtasks sequentially. We were officially ranked second in Phase 1, due to a submission error that resulted in phrase extraction F$_{1}$ of zero. Our correct submission achieved an average F$_{1}$ of 49.7\%, the best score among all participating teams. Table~\ref{phase1} shows our performance in Phase 1, and the intra-annotator agreement (IAA) on each subtask~\citep{dsouza2020graphing}.

We observe that, although the performance of our system on sentence classification is lower than human performance (57.27\% vs. 67.44\% F$_{1}$), using its own sentence predictions, our system outperforms human annotators on phrase recognition (46.41\% vs. 41.84\% F$_{1}$), and reaches comparable performance to human annotators on triple extraction. We also note that our system generally performs better in terms of recall than precision.
%, which suggests that additional post-processing could be helpful to improve the overall performance. \textcolor{orange}{Can this sentence be removed? It is not supported by later discussion.}

We were officially ranked first in both parts of Evaluation Phase 2. In Part 1, the participants were provided with the sentences labels to conduct phrase recognition and triple extraction sequentially; in Part 2, both the sentence labels and the phrase labels were provided to extract triples. We essentially followed our method in Phase 1 on phrase recognition and triple extraction, but made several attempts to improve the performance, which we discuss in Section~\ref{analysis}. Our results in both parts of the Phase 2 evaluation are shown in Table~\ref{phase2}. Compared to Phase 1 evaluation, we observe a significant improvement in phrase recognition (46.41\% vs. 78.57\% F$_{1}$) in Part 1 and in triple extraction (22.28\% to 43.44\% and 61.29\% F$_{1}$) when ground truth contribution sentences and phrases are provided. 

\begin{table*}[!htbp]
\centering
\resizebox{320pt}{!}{%
\begin{tabular}{|c|ccc|ccc|ccc|}
\hline
\multirow{2}{*}{} & \multicolumn{3}{c|}{Information Units} & \multicolumn{3}{c|}{Phrases} &
\multicolumn{3}{c|}{Triples} \\
\cline{2-10}
& F$_{1}$ & P & R & F$_{1}$ & P & R & F$_{1}$ & P & R \\ \hline
Part 1 & 82.49 & 76.84 & 89.02 & 78.57 & 76.86 & 80.35 & 43.44 & 45.06 & 41.94 \\
\hline
Part 2 & 82.49 & 76.84 & 89.02 & - & - & - & 61.29 & 65.19 & 57.82 \\
\hline
\end{tabular}
}
\caption{\label{phase2}
Performance in phrase and triple extraction (Evaluation Phase 2). Note that we focused only on triple extraction in Part 2, therefore the information unit extraction performance remains the same. 
}
\end{table*}

\begin{table*}[!htbp]
\centering
\small
\resizebox{\textwidth}{!}{%
\begin{tabular}{|C{1.2cm}|C{1.2cm}|C{1.2cm}|C{1.2cm}|C{1.2cm}|C{1.2cm}|C{1.2cm}|C{1.2cm}|C{1.2cm}|C{1.2cm}|C{1.2cm}|C{1.2cm}|C{1.2cm}|}
\hline
\textbf{Unit name} & Research problem & Approach & Model	& Code & Dataset & Experimental Setup & Hyperparameters & Baselines & Results & Tasks & Experiments & Ablation analysis \\ 
\hline
%\textbf{F$_{1}$} & 96.97 & 23.08 & 87.83 & 82.73 & 75.00 & 49.18 & 72.88 & 88.17 & 94.90 & 100.00 & 80.60 & 94.29 \\
%\hline
\textbf{F$_{1}$} & 94.64 & 24.14 & 86.22 & 87.50 & 80.00 & 58.29 & 72.61 & 91.45 & 94.65 & 90.48 & 83.16 & 90.68 \\
\hline
\end{tabular}
}
\caption{\label{units} Information unit classification performance. }
\end{table*}

\section{Performance Analysis} \label{analysis}
In this section, we analyze the performance of several components of our system and compare different schemes for entity representation and triple extraction. We also discuss some possible methods for improvement based on our shared task results and follow-up experiments.
\subsection{Contribution Sentence Classification}
We conducted ablation experiments to evaluate the effect of features for contribution sentence classification. Table~\ref{ablation} shows the model performance on the 10\% validation set when using all features, using either the title or the position features together with the sentence, and using the sentence only.

\begin{table}[!htbp]
\centering
%\small
\resizebox{\columnwidth}{!}{%
\begin{tabular}{|c|c|c|c|}
\hline
Settings & F$_{1}$ & P & R\\
\hline
Sentence + title + position & 65.11 & 63.96 & 66.30\\
\hline
Sentence + title & 63.87 & 61.00 & 67.03\\
\hline
Sentence + position & 52.28 & 46.38 & 59.89\\
\hline
Sentence only & 51.39 & 49.00 & 54.03\\
\hline
\end{tabular}
}
\caption{\label{ablation} Results of ablation experiments on contribution sentence classification task. }
\end{table}

We observe the title information significantly improves the performance, and the position features are also helpful, to a lesser extent. Combining the title and the position features gives the best performance on contribution sentence classification.

\subsection{Information Unit Classification}
In Evaluation Phase 2, the ground truth labels for contribution sentences increased the  performance of our base model on information unit classification from 72.93\% to 76.84\% F$_{1}$. To further improve our method, we ensembled 45 multi-class sentence classifiers by averaging their output (using \emph{bagging}), which increased the $F_1$ score to 78.65\%. Next, we improved our rules for distinguishing the special pairs (\textsc{Model} vs. \textsc{Approach} and \textsc{Experimental-setup} vs. \textsc{Hyperparameters}) by adjusting the lexical cues with more careful observation of the data, which results in our final performance (82.49\% F$_1$ in Table~\ref{phase2}).

\begin{figure}[!ht]
\begin{center} 
\includegraphics[width=\columnwidth]{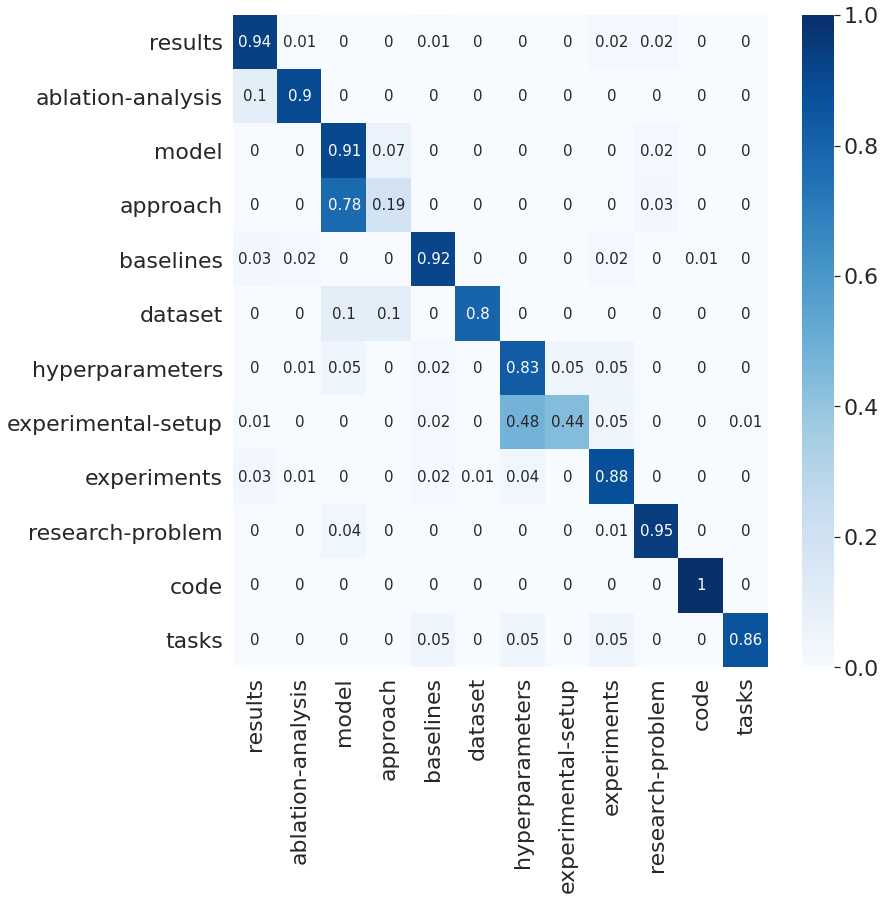}
\caption{\small Confusion Matrix\label{cm}} 
\end{center} 
\end{figure}

For further analysis, we evaluated the classification performance on each information unit, as shown in Table~\ref{units}. The related confusion matrix is shown in Fig.~\ref{cm}. We observe that severe confusion mainly occurs between \textsc{Model} vs. \textsc{Approach} and \textsc{Experimental-setup} vs. \textsc{Hyperparameters}, pairs that we grouped together in neural classification. This shows that while our sentence classification model has good accuracy, there is still much room for improvement in the rule-based differentiation of similar units.

\begin{table*}[!htbp]
\centering
\resizebox{1.4\columnwidth}{!}{%
\begin{tabular}{|c|c|c|c|c|c|c|}
\hline
\multirow{2}{*}{Settings} & \multicolumn{3}{c|}{Phrase extraction} & \multicolumn{3}{c|}{Phrase classification} \\
\cline{2-7}
& F$_{1}$ & P & R & F$_{1}$ & P & R \\
\hline
Specific BIO & 76.09 & 75.57 & 76.62 & 98.13 & 98.00 & 98.25 \\
\hline
Simple BIO & 77.13 & 76.33 & 77.95 & \multirow{2}{*}{98.27} & \multirow{2}{*}{98.70} & \multirow{2}{*}{97.85} \\
\cline{1-4}
Simple BIO + ensemble & 78.57 & 76.86 & 80.35 & & & \\
\hline
\end{tabular}
}
\caption{\label{phrase} Phrase extraction and classification performance. We take \emph{predicate} as the positive label to calculate the F$_{1}$ score for phrase classification.}
\end{table*}
The differentiation between \textsc{Model} and \textsc{Approach} is particularly challenging. While some papers aim at discussing an abstract idea and some focus on system implementation, most papers fall in the gray area between them. We also attempted neural classification on the abstracts to deal with this issue, but the result were not satisfactory. %We leave this for future work.

\subsection{Phrase extraction and classification}
\paragraph{Specific BIO VS. simple BIO}
Alternative to our method of using specific BIO tags to indicate phrase types (Subsection~\ref{ner}), we also used another scheme (``simple BIO''), in which we only used (B, I, O) tags to mark phrase boundaries.

With this scheme, we first trained a BERT-CRF model to extract the phrases, and then trained a binary BERT classifier to predict phrase types. The sentence along with the phrase marked by special tokens is fed into the BERT model for binary classification. The performance comparison of these schemes is shown in Table~\ref{phrase}. While both schemes are effective, simple BIO outperforms specific BIO in phrase extraction by a small margin, so we used this scheme in Evaluation Phase 2.

The difference may be due to the noise in phrase types. Specifically, there is a good number of gerund phrases, on which the predicate-term differentiation is challenging. Moreover, in some cases, a verb phrase is used as a term to form triples. Combined with the relatively low intra-annotator agreement, these observations suggest that uncertainty and noise in the data affects the performance of the models.  
%The uncertainty in phrase types more or less reduces the annotation quality which affects the performance of the model trained on these tags. 
Note that the specific BIO scheme eliminates the need for a separate phrase classification model, making it preferable when the training and inference speed is a concern.
\paragraph{Error analysis and improvement}
We investigated the wrong predictions of our phrase extraction model, and found that most errors are due to boundary detection issues. For example, in one sentence, the model predicts \emph{all layers of representation} as a phrase, while \emph{all layers}, \emph{of}, \emph{representations} are annotated as three separate phrases. The opposite situation also occurs, when the model predicts a single unit as separate phrases. Another type of boundary error occurs when the model cannot predict correctly whether to include a non-core phrase element, like an adverb, in the phrase or not (e.g., it predicts \emph{see that} whereas the annotated phrase is \emph{also see that}). We believe that a relaxed boundary match evaluation can be considered for this task.

We attribute these errors to the uncertainty in semantic granularity, and attempted to alleviate the problem by ensembling. We get 12 bootstrap samples from the training data, and on each sample, we train the model and save its snapshot after each epoch from the 3th epoch to the 10th epoch, to get a total of 96 submodels. To aggregate their predictions, we extract a phrase in a sentence only if it is predicted by more than N submodels, where N is a hyperparameter around 48. We present the result in Table~\ref{phrase} for comparison. We observe that ensembling noticeably improved phrase extraction (from 77.13\% to 78.57\% F$_{1}$).

\subsection{Triple extraction}
\paragraph{Triple vs. pairwise classification}
 In addition to triple classification method (Subsection~\ref{typea}) to extract type A triples,  we also used pairwise classification for this task. In this scheme, we considered every (subject, predicate, object) triple as a composition of two (predicate, term) pairs, or ``candidate pairs'', and used a neural model to predict whether the two phrases in the pair are associated. 
 %To train this model, we labeled the pairs that exist in an annotated phrase as positive, and labeled the rest as negative. 
 After prediction, we reconstructed triples from the predicted pairs using rules. If a predicate is predicted to be associated with two terms, we combine them into a triple while preserving the order of the two terms in the sentence (subject first). If one predicate is associated with more than two terms, we only extract the triples in which the predicate is located between the two terms in the sentence. With only a few exceptions, we confirmed the effectiveness of these reconstruction rules; in other words, the performance of the pairwise scheme depends mainly on the classification accuracy on candidate pairs. 

We compared the performance of the two schemes for type A triple extraction on the 10\% validation set. We also attempted to address the imbalance of class labels resulting from both schemes by downsampling and class weight adjustment.

\begin{table}[!ht]%[!htbp]
\centering
\resizebox{\columnwidth}{!}{%
\begin{tabular}{|c|c|c|c|c|}
\hline
\multicolumn{2}{|c|}{Settings} & F$_{1}$ & P & R\\
\hline
No & Pair & 91.33 & 91.23 & 91.43 \\
\cline{2-5}
adjustment & Triple & 75.95 & 70.58 & 82.20 \\
\hline
\multirow{2}{*}{Downsampling} & Pair & 91.31 & 89.09 & 93.63 \\
\cline{2-5}
& Triple & 80.04 & 79.43 & 80.66 \\
\hline
{Class} & Pair & 91.30 & 88.93 & 93.79 \\
\cline{2-5}
weight & Triple & \textbf{80.37} & 81.35 & 79.42 \\
\hline
\end{tabular}
}
\caption{\label{pairwise} Performance of the pairwise classification scheme.}
\end{table}

\begin{table}[!ht]%[!htbp]
\centering
\resizebox{0.85\columnwidth}{!}{%
\begin{tabular}{|c|c|c|c|}
\hline
& F1 & P & R\\
\hline
No adjustment & \textbf{87.54} & 85.93 & 89.22 \\
\hline
Downsampling & 75.59 & 62.32 & 96.04 \\
\hline
Class weight & 83.35 & 74.94 & 93.89 \\
\hline
\end{tabular}
}
\caption{\label{triple} Performance of the triple classification scheme. }
\end{table}

In the pairwise classification scheme (Table~\ref{pairwise}), there is a 11\% drop in the F1 score from the candidate pair classification to triple prediction, which is not unexpected as the model needs to correctly classify \emph{both} of the candidate pairs to correctly predict a triple.  

%While we manually validated that triple reconstruction rules are accurate, when the We have validated that the reconstruction rules are very accurate, but the performance is decreased after the reconstruction process. This is because in order to correctly predict a triple, the model needs to correctly classify both of the two candidate pairs. Thus the performance decrease is inevitable from a probabilistic viewpoint.

Table~\ref{triple} shows the performance of the triple classification scheme, which achieves better performance compared to the pairwise classification scheme (87.54\% vs. 80.37\% F$_{1}$). We also observed that the best performance was obtained without dealing with the imbalanced data. It seems that despite constituting a small portion of the dataset (9.7\%) , the number of the positive samples is large enough for the model to learn useful patterns.
\paragraph{Type-specific performance}
We also evaluated our deep learning methods for the extraction of the four types of triples, as shown in Table~\ref{types}.
\begin{table}[!htbp]
\centering
\resizebox{0.7\columnwidth}{!}{%
\begin{tabular}{|c|c|c|c|}
\hline
Type & F1 & P & R\\
\hline
A & 87.54 & 85.93 & 89.22 \\
\hline
B & 55.56 & 88.24 & 40.54 \\
\hline
C & 83.33 & 77.96 & 89.51 \\
\hline
D & 75.86 & 78.11 & 73.74 \\
\hline
\end{tabular}
}
\caption{\label{types} Performance of triple extraction on each type. }
\end{table}

Whereas our models for Type A, C, and D perform generally well, our model for Type B is far less accurate. Type B is a little special among the four types in that it requires the prediction of relation types. The type \texttt{has} is more difficult to predict than \texttt{name}, because the sentence often lacks semantic clues about the belonging or inclusion relationship between the two terms. A plausible idea is to incorporate \emph{has} into the input, but it is difficult to do so without breaking the grammatical integrity of the sentence. We leave this improvement for future work.

\paragraph{Coordination in triple extraction}
A common problem we observed in our triple extraction models is the failure to account for coordination between terms.  
%Despite a good overall performance, our triple extraction method fails to capture some patterns, leading to inconsistent predictions. 
Example~\ref{ex3} shows a sentence with the terms in bold, and the two type C triples associating them. Our model only extracts the first triple, and misses the second. 
\begin{expl}
\emph{The MoE consists of a \textbf{number of experts}, each a simple feed - forward neural network, and a \textbf{trainable gating network} which selects a sparse combination of the experts to process each input.}\\
(\textsc{Approach}, \texttt{consists of}, \emph{number of experts})\\
(\textsc{Approach}, \texttt{consists of}, \emph{trainable gating network})
\label{ex3}
\end{expl}

%Our method only considers the first term in each sentence as the object for a type C triple, so it only extracts the first triple, and omits the second. This is not reasonable, since the two triples only differ in objects, and the two objects are coordinated.
We attempted to address this issue in post-processing, and used Stanza dependency parser~\citep{qi-etal-2020-stanza} to detect coordination of words in phrases. If one phrase is used in a triple, we generated a parallel triple by replacing the term with the other. While this method improves recall (from 57.57\% to 58.41\%), it also led to precision errors (from 65.15\% to 61.77\%), its overall effect being negative (from 61.13\% to 60.04\% F$_{1}$). We plan to refine this approach in future work.
\section{Conclusion}
We developed a system to generate structured representations of research contributions described in NLP publications in a manner compatible with the ORKG framework, achieving the top performance in the NCG shared task. We combined a cascade of state-of-the-art BERT-based classification and sequence labeling models with rule-based methods.  In particular, we proposed a novel approach for triple extraction, where we tackled triples with different characteristics using different relation classification methods. We also explored various alternatives to the components in our end-to-end system to analyze the contribution of individual components. 

In future work, we plan to improve the differentiation of similar units (e.g., \textsc{Model} vs. \textsc{Approach}), improve the extraction of type B triples, and address coordinated triples more thoroughly. We did not attempt to extract approximately 6\% of the triples that did not fit in our classification~(Table \ref{tripletypes}). These often involve nested information units, and we also hope to explore them in more depth in future work.
%We noticed that there are 6% triples outside of these categories, for which we haven't found reliable patterns to extract. A large portion of them contain a nested info unit (e.g. Results, Experimental setup, or Tasks nested under the Experiment unit) as the subject. We hope to explore them deeper in future work. (edited) 
%We also would like to improve the extra as well as   Our results are pro triples using differenwrepresentations and explored  Our most important contribution is a combination of rule-based and deep learning-based methods that achieves high performance on the triple extraction task. We proposed a novel way to classify triples directly, which is more accurate on this task than the common relation classification method. We also evaluated different components of our system and found some limitations that we hope to improve in future work. 

\bibliographystyle{acl_natbib}
\bibliography{anthology,acl2021}

%\appendix

\end{document}